\documentclass[10pt,twocolumn,letterpaper]{article}

\PassOptionsToPackage{table}{xcolor}
\usepackage[pagenumbers]{cvpr}
\usepackage{multirow}
\usepackage{url}
\usepackage{amssymb}
\usepackage{pifont}
\usepackage{cuted}
\usepackage{amsmath}
\usepackage{graphicx}
\usepackage{booktabs}
\usepackage[table]{xcolor}
\usepackage{comment}
\usepackage{wrapfig}
\usepackage{caption}
\usepackage{algorithm}
\usepackage{algpseudocode}

\newcommand{\suppname}{the appendix}
\newcommand{\webname}{project webpage}
\definecolor{cvprblue}{rgb}{0.21,0.49,0.74}
\definecolor{iccvblue}{rgb}{0.21,0.49,0.74}
\usepackage[pagebackref,breaklinks,colorlinks,allcolors=cvprblue]{hyperref}

\def\confName{}
\def\confYear{}

\title{\raisebox{-0.25em}{\includegraphics[height=1.25em]{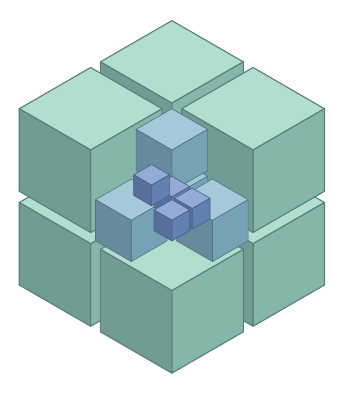}}\hspace{0.25em}JoVox: Joint Hierarchical Flow Matching for Lightweight Voxel Generation}

\author{Xiangjun Tang \qquad Jianqi Chen \qquad Peter Wonka\\
King Abdullah University of Science and Technology (KAUST)\\
{\tt\small xiangjun.tang@outlook.com \quad windvchen@gmail.com \quad pwonka@gmail.com}
}

\begin{document}
\maketitle

\begin{strip}
    \centering
    \vspace{-4em}
    \includegraphics[width=1.0\linewidth]{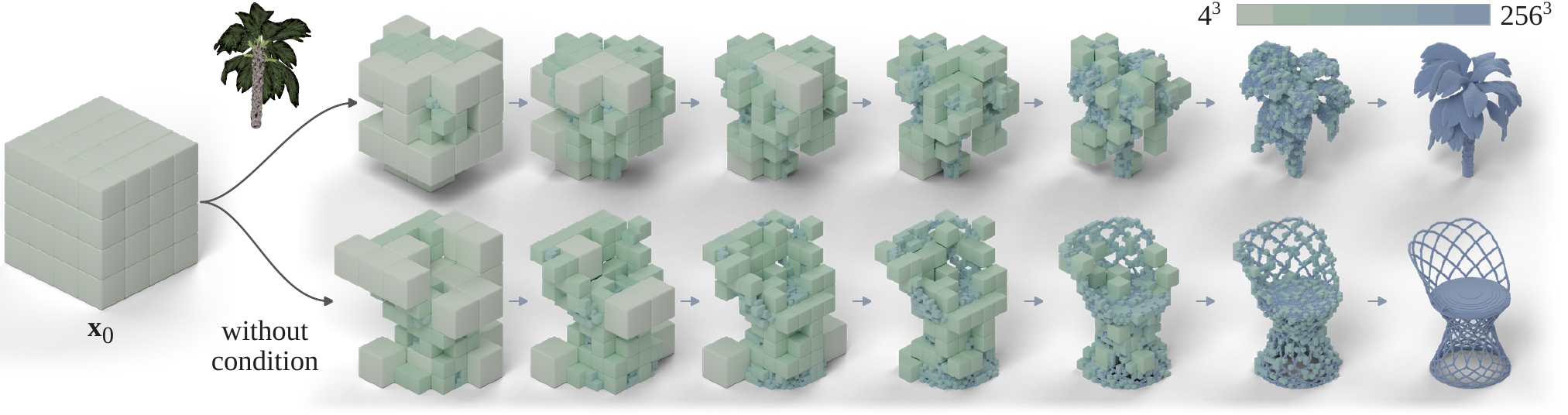}
    \vspace{-2em}
    \captionof{figure}{From the same $4^3$ grid $\mathbf{x}_0$, we
    generate voxels conditioned on the input image (top) and without a condition (bottom). We learn
    the \emph{joint distribution} of the voxels at all resolutions, instead of a separate one for each resolution.
    Finer resolutions thus start before coarser ones are complete, and every intermediate state
    mixes several resolutions, each shown in its own color.}
    \label{fig:teaser}
\end{strip}
\begin{abstract}
Generating high-fidelity voxels while keeping their lightweight nature remains an open
problem in computer vision. Current generators are either high-fidelity but computationally
heavy, or lightweight but of lower fidelity. Hierarchical methods keep the lightweight nature
of voxels. However, their coarse levels, which decide where all finer cells can appear, are
generated with the least information about the shape and are never revisited, so they are
often plausible but wrong. Our key insight is to let
finer cells take part in deciding coarse cells. To this end, we propose JoVox, a discrete flow matching framework that learns the joint distribution of all levels of the
voxel hierarchy, instead of a separate distribution for each
level, so that coarse cells can be predicted with finer cells already generated nearby. Within
this framework, we further introduce a two-level prediction, which outputs the cells two levels
finer at once instead of one level finer, halving the number of levels that the network processes and
yielding voxels of higher fidelity. We also introduce an unmasking strategy that revisits early decisions once finer
cells are generated and corrects inconsistencies in masked diffusion models. Our framework outperforms the baselines on both unconditional and
image-conditioned voxel generation. Compared with the current state-of-the-art TRELLIS.2, we
achieve $8\times$ higher training throughput, $1/15$ of the FLOPs in a forward pass, train on $4$
vs.\ $32$ GPUs, and obtain higher quality with an $11\%$ lower Chamfer distance.
\end{abstract}

\section{Introduction}
\label{sec:intro}

Voxel generation is an important problem in computer vision. Voxels record where a 3D
shape lies in space, giving a lightweight and widely used 3D representation. For example, existing models can turn voxels into fine-grained
3D shapes~\cite{xiang2025structured,xiang2026native}, keeping the structure that the voxels define. Yet
generating high-fidelity voxels while keeping this lightweight nature remains an open
problem.

Existing methods address this problem in two ways. Latent
methods~\cite{xiang2025structured,xiang2026native} achieve high fidelity by compressing the
voxel grid with an autoencoder and generating the latent grid with a generator. The autoencoder is an extra
model to train, and the generator still processes every cell of the latent grid, including
those in empty space, so the computational cost remains high. Hierarchical
methods~\cite{ibing2023octree,xiong2025octfusion,wei2025octgpt} instead generate a coarse
grid first and refine its occupied cells into finer grids, level by level, which keeps the
lightweight nature of voxels. However, the coarse levels, which decide where all finer cells
can appear, are generated with the least information about the shape and are never revisited.
They are thus often plausible but wrong, and the finer levels inherit these errors.

Our key insight is to let finer cells take part in deciding coarse cells. To this end, we
propose JoVox, a discrete flow matching framework that learns the joint distribution of all
levels of the hierarchy, instead of a separate distribution for each level as in
hierarchical methods~\cite{ibing2023octree,xiong2025octfusion,wei2025octgpt}. Such a
distribution is hard to model, since the structure of the hierarchy varies across shapes, with
finer cells only inside occupied cells. We model it with masked discrete flow
matching~\cite{lipman2024flow,gat2024discrete}, which unmasks a few cells at each step, at any
level of the hierarchy, and grows the hierarchy as cells are unmasked. A coarse cell that is
still masked can then be predicted with the finer cells already unmasked nearby. Our framework
needs no autoencoder and processes only the cells inside occupied cells, which keeps generation
lightweight.

Within this framework, we further introduce two-level prediction and an unmasking strategy.
Two-level prediction outputs the cells two levels finer at once, instead of one level finer as in
hierarchical methods~\cite{ibing2023octree,deng2025efficient,wei2025octgpt}, and thus halves
the number of levels that the network processes. We also find that two-level prediction yields
voxels of higher fidelity (Section~\ref{sec:exp-ablation}). Our unmasking strategy remasks
the oldest unmasked cells, which were decided with the fewest finer cells around them, so that
they are revisited once finer cells are generated. This also corrects the inconsistencies of
cells unmasked in the same step~\cite{liu2025copula,besnier2025halton}.

We evaluate our method on both unconditional and image-conditioned voxel generation, where
it outperforms current SOTA. In particular, our framework makes training far cheaper
than that of TRELLIS.2~\cite{xiang2026native}. With the same backbone, our
framework needs significantly less GPU memory and has $8\times$ higher
training throughput (Figure~\ref{fig:efficiency}), while its voxels remain of higher fidelity,
with an $11\%$ lower Chamfer distance.

In summary, our contributions are as follows.
\begin{itemize}
  \item JoVox, a discrete flow matching framework that learns the joint hierarchical
  distribution of voxels while keeping generation lightweight.
  \item Two-level prediction, which outputs the cells two levels finer instead of one level
  finer, halving the number of levels that the network processes and yielding voxels of higher
  fidelity, and an unmasking strategy that revisits early decisions
  once finer cells are generated and corrects inconsistencies in masked diffusion models~\cite{liu2025copula,besnier2025halton}.
  \item Compared with the current state-of-the-art TRELLIS.2, we achieve $8\times$ higher
  training throughput, $1/15$ of the FLOPs in a forward pass, train on $4$ vs.\ $32$ GPUs, and
  obtain higher quality with an $11\%$ lower Chamfer distance.
\end{itemize}

\section{Related work}
\label{sec:related}

\noindent\textbf{Voxel generation.} Voxels are generated as the first stage of many 3D
generators~\cite{xiang2025structured,xiang2026native,xiang2026dvd,wu2026direct3d}, or obtained
by voxelizing the meshes of 3D generators that produce meshes
directly~\cite{tochilkin2024triposr,li2025triposg}. Methods that generate voxels fall into
two groups, dense methods and hierarchical methods.

Dense methods generate every cell of a full grid.
TRELLIS~\cite{xiang2025structured,xiang2026native} compresses the voxels into a compact
latent grid with an autoencoder and generates this latent grid, which reaches high
fidelity at the cost of training an additional autoencoder. LAS-Diffusion~\cite{zheng2023lasdiffusion} and DVD~\cite{xiang2026dvd} instead
generate the occupancy grid directly. LAS-Diffusion generates the occupancy with continuous
diffusion and thresholds the result. DVD treats the occupancy as discrete data and
generates it with discrete diffusion.
Since all these methods generate every cell, including those in empty space, their cost
grows with the cube of the resolution.

Hierarchical methods generate the voxels level by level and refine only occupied cells,
which skips empty space. XCube~\cite{ren2024xcube} generates a sparse voxel hierarchy, with
its own latent diffusion model and autoencoder at each level.
Nexus~\cite{wang2026nexus} and OctFusion~\cite{xiong2025octfusion} generate the octree with
continuous diffusion and threshold the result to decide which cells to refine. Octree
Transformer~\cite{ibing2023octree} and OAT~\cite{deng2025efficient} generate the octree
autoregressively, one cell at a time, so the number of forward passes grows with the number
of cells. OctGPT~\cite{wei2025octgpt} generates many cells of a level in parallel by
masking. However, it decides the finer cells inside a cell independently and needs many
steps at every level.
In all these methods, a coarse level is decided from the coarser levels alone. Such a
level is only roughly plausible.

We instead learn the joint distribution of all levels, so that
finer cells take part in deciding coarse cells. Each prediction also outputs the cells two levels finer, instead of
one level finer. Like the hierarchical methods above, we refine only occupied cells. We
generate the occupancy itself, without an autoencoder. Table~\ref{tab:related} summarizes
these differences.

\begin{table}[t]
\centering
\caption{Comparison of voxel generation methods.}
\vspace{-0.5em}
\label{tab:related}
\setlength{\tabcolsep}{4pt}
\resizebox{\columnwidth}{!}{%
\begin{tabular}{lcccc}
\toprule
Method & Hierarchy & No autoencoder & Parallel & Joint levels \\
\midrule
TRELLIS, TRELLIS.2 & \ding{55} & \ding{55} & \checkmark & -- \\
LAS-Diffusion & \ding{55} & \checkmark & \checkmark & -- \\
DVD & \ding{55} & \checkmark & \checkmark & -- \\
\midrule
XCube & \checkmark & \ding{55} & \checkmark & \ding{55} \\
Nexus, OctFusion & \checkmark & \checkmark & \checkmark & \ding{55} \\
Octree Transformer, OAT & \checkmark & \checkmark & \ding{55} & \ding{55} \\
OctGPT & \checkmark & \checkmark & \checkmark & \ding{55} \\
\midrule
Ours & \checkmark & \checkmark & \checkmark & \checkmark \\
\bottomrule
\vspace{-3em}
\end{tabular}%
}
\end{table}

\noindent\textbf{Discrete diffusion and flow matching.} Masked discrete diffusion
models~\cite{austin2021structured,sahoo2024simple,lou2023discrete,chang2022maskgit} and
discrete flow matching~\cite{lipman2024flow,gat2024discrete} generate discrete data by
unmasking a masked sequence over several steps, and have been applied to text and image
tokens. These data have a fixed size, whereas a voxel hierarchy grows as its cells are
unmasked. We extend masked discrete flow matching to such a hierarchy.

When several tokens are unmasked in one step, these models sample them independently, which
can make them inconsistent with each other. Some works model the dependence among the tokens
unmasked together~\cite{hayakawa2024distillation,liu2025copula,xu2025energy}. Other works
choose which tokens to unmask
together~\cite{wu2026fast,besnier2025halton,luxembourg2025plan,ringel2026dependency}, or
remask unmasked tokens so that they can be revised~\cite{wang2026remasking}. Our unmasking
strategy also remasks, but chooses the oldest unmasked cells, which were decided with the
fewest finer cells around them.

\section{Discrete flow matching}
\label{sec:prelim}

Discrete flow matching~\cite{lipman2024flow,gat2024discrete} transports samples of a source distribution
into samples of a data distribution, along a path indexed by $t\in[0,1]$. A data sample
is a tensor $\mathbf{x}=(x^1,\dots,x^n)$ with $n$ entries from a finite vocabulary
$\mathcal{T}$. The $i$-th entry of $\mathbf{x}$ is $x^i$. Throughout, $\mathbf{x}_1$ denotes a
sample of the data distribution, and $\mathbf{x}_0$ a sample of the source
distribution, which is a modeling choice. $\mathbf{x}_t$ denotes the tensor
at $t$ on the path between $\mathbf{x}_0$ and $\mathbf{x}_1$.

A common modeling choice of the source distribution is the masked
source~\cite{austin2021structured,sahoo2024simple}. Instead of
sampling from a probability distribution, the masked source is a fully masked tensor
$\mathbf{x}_0=(\varnothing,\dots,\varnothing)$, where $\varnothing\notin\mathcal{T}$ is
a mask symbol.

\textbf{Masking.} The path from the data to the source is thereby a masking process.
Specifically, given $\mathbf{x}_1$, each entry independently keeps its value with
probability $t$ and is masked otherwise. The distribution of $\mathbf{x}_t$ given
$\mathbf{x}_1$ is
\begin{equation}
p_{t|1}(\mathbf{x}_t\mid \mathbf{x}_1)=\prod_{i=1}^{n}\Big[
t\,\delta(x_t^i,x_1^i)+(1-t)\,\delta(x_t^i,\varnothing)\Big],
\label{eq:path}
\end{equation}
with $\delta(a,b)=1$ if $a=b$ and $\delta(a,b)=0$ otherwise. An entry whose
value is in $\mathcal{T}$ is called unmasked.

\textbf{Unmasking.} A flow matching model generates data by unmasking. One forward pass
of the model, a network with parameters $\theta$, returns probabilities over $\mathcal{T}$ for every entry,
\begin{equation}
p_\theta(x_1^i\mid\mathbf{x}_t,t),\qquad i=1,\dots,n,
\label{eq:model}
\end{equation}
each approximating the distribution of $x_1^i$ given $\mathbf{x}_t$. For the currently masked entries, these probabilities are predicted considering all already unmasked entries of
$\mathbf{x}_t$. At each step of generation, a random subset of the masked entries is unmasked, each with a
value sampled from its predicted probabilities.

\textbf{Loss.} The model is fitted by minimizing the loss of discrete flow
matching~\cite{lipman2024flow,gat2024discrete}. This loss uses the predicted probabilities of both masked
and unmasked entries. For the masked path, the loss of entry $i$ is
\begin{equation}
\ell^i=\begin{cases}
-\log p_\theta(x_1^i\mid\mathbf{x}_t,t) & \text{if } x_t^i=\varnothing,\\
1-p_\theta(x_1^i\mid\mathbf{x}_t,t) & \text{otherwise}.
\end{cases}
\label{eq:entryloss}
\end{equation}
The first case is the cross-entropy on a masked entry. The second penalizes a
low probability of the unmasked value. The model minimizes
\begin{equation}
\mathcal{L}=\int_0^1\frac{1}{1-t}\,\mathbb{E}\Big[\textstyle\sum_{i}\ell^i\Big]\,\mathrm{d}t.
\label{eq:elbo}
\end{equation}

\section{Method}
\label{sec:method}

Given an optional condition $c$, such as an image, our task is to generate the occupancy of a
shape. We build upon masked discrete flow matching (Section~\ref{sec:prelim}) to model the
joint distribution of all levels of the voxel hierarchy.
Section~\ref{sec:process} defines the hierarchy and models its levels jointly, and
Section~\ref{sec:twolevel} introduces two-level prediction.
Sections~\ref{sec:network} and~\ref{sec:training} describe the network and the
training, and Section~\ref{sec:sampling} introduces the unmasking strategy for inference.

\begin{figure*}[t]
  \centering
  \vspace{-1em}
  \includegraphics[width=\linewidth]{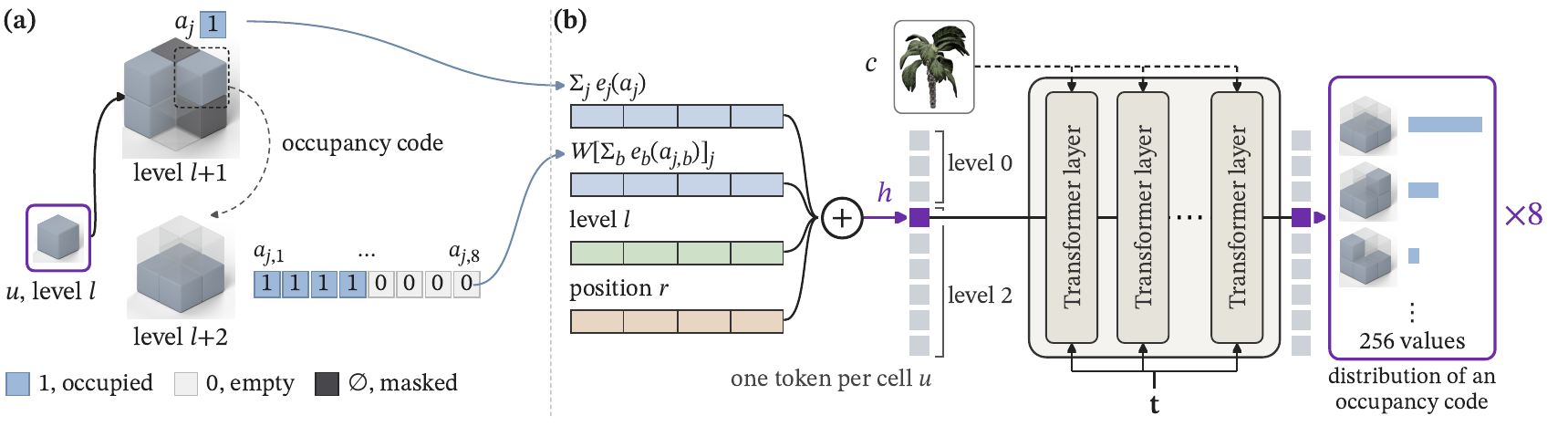}
  \caption{(a)~Two-level prediction. For each cell $u$ at level $l$, the network
  predicts the occupancy codes of the eight children of $u$ at level $l+1$. The occupancy code of the
  $j$-th child consists of the values $a_{j,1},\dots,a_{j,8}$ of its eight children at level
  $l+2$, where a value is $1$ for an occupied cell, $0$ for an empty one and $\varnothing$ for a
  cell whose occupancy is unknown. When this occupancy code is masked, the value $a_j$ of the $j$-th
  child and $a_{j,1},\dots,a_{j,8}$ are all $\varnothing$.
  (b)~The network. Each cell $u$ at level $0$ and each occupied cell $u$ at even levels has one token.
  The token $h$ embeds the values $a_j$ and $a_{j,b}$, the level $l$ and the position $r$. A transformer processes the tokens of all levels together, conditioned on $c$ and on
  the fractions $\mathbf{t}$ of unmasked occupancy codes, and outputs for $u$ eight distributions
  over the $256$ values of an occupancy code.}
  \label{fig:pipeline}
\end{figure*}

\subsection{Joint hierarchical modeling}
\label{sec:process}

A shape is represented by its occupied cells at the levels
$l=0,\dots,L$. Level $l$ is a grid of $(4\cdot2^{l})^3$ cells, and a cell is occupied if the
shape surface passes through it.
Each cell at level $l<L$ contains eight cells at level $l+1$, which
we call its children. As in an octree, a cell is identified by its level $l$ and its position in the grid of level $l$. We denote
by $O_l$ the set of occupied cells at level $l$ and sample $O_0,\dots,O_L$ from
$p(O_0,\dots,O_L\mid c)$.

Level-by-level generation factorizes this distribution into $p(O_0\mid c)\prod_{l=1}^{L}p(O_l\mid O_{l-1},c)$, so
each level is decided given only the coarser levels. We instead learn the joint distribution of
all levels with the masked discrete flow matching of Section~\ref{sec:prelim}, so that a
coarse cell can be predicted given finer cells that are already generated.

\textbf{Occupancies.}
Without loss of generality, we start with a straightforward choice of the entries $x^i$ of Section~\ref{sec:prelim} for hierarchical modeling. We define the entry $x^i \in \mathbf{x}_1$ as the occupancy of a cell. The occupancy is $1$ if the cell is occupied and $0$ otherwise,
so that $x^i\in\mathcal{T}=\{0,1\}$. During discrete flow matching, an occupancy can also be masked, and a masked occupancy is the mask symbol $\varnothing$ of Section~\ref{sec:prelim}.

\textbf{Masking.} The data sample $\mathbf{x}_1$ contains the occupancies of the cells at level $0$ and of the children of occupied cells, and it determines
$O_0,\dots,O_L$. As in Section~\ref{sec:prelim}, we obtain an intermediate state by masking some
occupancies of $\mathbf{x}_1$. Masking the occupancy of a cell also removes all cells inside it, since their occupancies would reveal the masked occupancy. We mask the levels $J=\{0,\dots,L\}$ in order, from the coarsest to the finest. At level $l$,
each occupancy keeps its value with probability $t_l$ and is masked otherwise.
The resulting intermediate state $\mathbf{x}_{\mathbf{t}}$, with $\mathbf{t}=(t_l)_{l\in J}$, thus contains the occupancies of $\mathbf{x}_1$ whose cells are not removed, and each of these occupancies keeps its value or is masked. $\mathbf{x}_{\mathbf{t}}$
follows
\begin{equation}
p_{\mathbf{t}|1}(\mathbf{x}_{\mathbf{t}}\mid\mathbf{x}_1)=\prod_{l\in J}\,
\prod_{i\in \mathcal{C}_l(\mathbf{x}_{\mathbf{t}})}
\Big[t_l\,\delta(x_{t_l}^i,x_1^i)+(1-t_l)\,\delta(x_{t_l}^i,\varnothing)\Big],
\label{eq:hpath}
\end{equation}
where $\mathcal{C}_l(\mathbf{x}_{\mathbf{t}})$ is the set of occupancies at level $l$ in
$\mathbf{x}_{\mathbf{t}}$. The number of occupancies thus depends on which cells are known to be occupied. It differs between
shapes, and even for the same shape it changes as the shape is built during discrete flow matching.

\textbf{Unmasking.} Given $\mathbf{x}_{\mathbf{t}}$, $\mathbf{t}$ and $c$, the model processes each cell whose occupancy is in $\mathbf{x}_{\mathbf{t}}$ and returns
probabilities over $\mathcal{T}$ for its occupancy,
\begin{equation}
p_\theta(x_1^i\mid\mathbf{x}_{\mathbf{t}},\mathbf{t},c),\qquad
i\in \mathcal{C}_l(\mathbf{x}_{\mathbf{t}}),\ l\in J.
\label{eq:hmodel}
\end{equation}
Generation starts from $\mathbf{x}_0$. At each step, we choose masked occupancies at any
level at random (Section~\ref{sec:sampling}) and unmask them by sampling their values from the probabilities of Equation~\eqref{eq:hmodel}. Once a cell at level $l<L$ is known to
be occupied, we add masked occupancies of its children to $\mathbf{x}_{\mathbf{t}}$, so the hierarchy grows as occupancies are unmasked. Since occupancies at any level can be unmasked, some occupancies at a
coarse level can still be masked when occupancies at finer levels in nearby cells are already
unmasked.

\subsection{Two-level prediction}
\label{sec:twolevel}

Hierarchical methods such as Octree Transformer~\cite{ibing2023octree} define the \textit{occupancy code} of a cell as the occupancy of its eight children, which has 8 bits. They process a cell to predict its occupancy code, one level finer, instead of its own occupancy as in Section~\ref{sec:process}. We call this single-level prediction. In our framework, single-level prediction uses the formulation of Section~\ref{sec:process} with occupancy codes in place of occupancies. The entries $x^i\in\mathcal{T}=\{0,1\}^8$ are the occupancy codes at the levels $J=\{0,\dots,L-1\}$. The data sample $\mathbf{x}_1$ contains the occupancy codes of the cells at level $0$ and of the occupied cells at the levels $1,\dots,L-1$. We mask the occupancy codes as in Section~\ref{sec:process} to obtain $\mathbf{x}_{\mathbf{t}}$, where $\mathcal{C}_l(\mathbf{x}_{\mathbf{t}})$ is now the set of occupancy codes at level $l$ in $\mathbf{x}_{\mathbf{t}}$. Once a cell at the levels $1,\dots,L-1$ is known to be occupied, we add its masked occupancy code to $\mathbf{x}_{\mathbf{t}}$.

In two-level prediction, the model instead processes a cell $u$ at level
$l\le L-2$ to predict the occupancy of its $4\times4\times4$ cells two levels finer, at level $l+2$
(Figure~\ref{fig:pipeline}~(a)). One possible way is to use the occupancy of these cells as one entry, but such an entry would
take $2^{64}$ values, which is prohibitively large. We therefore split the occupancy of these cells into eight 8-bit occupancy codes, one for each child of $u$, each giving the occupancy of the eight cells at level $l+2$ inside that child. For a cell $u$, the model thus takes these eight occupancy codes, masked or unmasked, as input and outputs eight distributions of Equation~\eqref{eq:hmodel}, each over the $2^8$ values of an occupancy code, instead of one distribution over $2^{64}$ values.

Specifically, we take $L$ to be even and let the cells $u$ lie at the even levels $0,2,\dots,L-2$, so that the entries are the occupancy codes at the odd levels $J=\{1,3,\dots,L-1\}$. The data sample $\mathbf{x}_1$ contains the 8-bit occupancy codes of the eight children of every cell $u$ at level $0$ and of every occupied cell $u$ at the levels $2,\dots,L-2$. We mask the occupancy codes as in single-level prediction to obtain $\mathbf{x}_{\mathbf{t}}$. Once a cell $u$ at the levels $2,\dots,L-2$ is known to be occupied, we add the masked occupancy codes of its eight children to $\mathbf{x}_{\mathbf{t}}$. The model thus processes each cell $u$ at level $0$ and each cell $u$ known to be occupied.
Two-level prediction thus processes cells at only half of the levels of single-level prediction, so training is faster
(Section~\ref{sec:exp-ablation}).

\subsection{Network}
\label{sec:network}

The network is a transformer that takes one token for each cell $u$ of Section~\ref{sec:twolevel}, as
illustrated in Figure~\ref{fig:pipeline}~(b).

As in Figure~\ref{fig:pipeline}~(a), we number the children of a cell $1,\dots,8$ in a fixed
order. We write $a_j$ for the value of the $j$-th child of $u$ and $a_{j,b}$ for the value of the
$b$-th child of this child. A value is $1$ if the cell is occupied, $0$ if it is empty, and
$\varnothing$ if the occupancy code of the $j$-th child is masked. The token $h$ of $u$ embeds these
values, the level $l$ and the position $r$ of $u$ as
\begin{equation}
h = \sum_{j=1}^{8} e_j(a_j) + W\Big[\sum_{b=1}^{8} e_b(a_{j,b})\Big]_{j=1}^{8}
+ \mathrm{emb}_{\text{level}}(l) + \mathrm{emb}_{\text{pos}}(r),
\label{eq:token}
\end{equation}
where $e_m$ is a
learned embedding of the values $\{0,1,\varnothing\}$ for the $m$-th child of a cell, used both
for the children of $u$, as $e_j$, and for the cells inside each child, as $e_b$. The first term embeds the eight children. The second sums
the embeddings of the eight cells inside each child. The eight sums are concatenated in the
order of the children, written $[\cdot]_{j=1}^{8}$, and a learned linear map $W$ maps them to
one vector.
The level embedding is one learned vector per level. The point $r\in[0,1)^3$ is the center of
$u$ in a frame shared by all levels. $\mathrm{emb}_{\text{pos}}(r)$ is the point embedding of
3DShape2VecSet~\cite{biao2023VecSet}.

The transformer has the same architecture as the voxel stage of
TRELLIS.2~\cite{xiang2026native}. It processes the tokens of all levels in one sequence, with
full self-attention and 3D rotary position embeddings of $r$. When $c$ is an image, we feed its
DINOv3 features~\cite{simeoni2025dinov3} to every block through cross-attention. We embed each $t_l$ with its own time embedding
$\mathrm{emb}_t^{\,l}$, and condition all tokens on the sum $\sum_{l\in J}\mathrm{emb}_t^{\,l}(t_l)$
through adaptive layer normalization in every block.

\subsection{Training}
\label{sec:training}

During training, we first draw $t_l$ for each level $l\in J$ and then draw
$\mathbf{x}_{\mathbf{t}}$ from Equation~\eqref{eq:hpath}. We draw the $t_l$ of different
levels independently, so the network is trained on any combination of fractions across levels. The
intermediate states $\mathbf{x}_{\mathbf{t}}$ produced by any unmasking strategy therefore lie
within the training distribution. To train the network more where its predictions are least
certain, we draw small $t_l$ more often, following the distribution given in
Section~\ref{sec:impl} in \suppname{}.

Finer levels have far more occupancy codes than coarser ones, so a loss summed over all occupancy codes
would be dominated by the finer levels. We therefore first average the loss over the
occupancy codes of each level, and then average over the levels. With
$J(\mathbf{x}_{\mathbf{t}})$ the set of levels that have occupancy codes in
$\mathbf{x}_{\mathbf{t}}$, the loss is
\begin{equation}
\mathcal{L}_{\text{ours}}=\mathbb{E}\Big[\frac{1}{|J(\mathbf{x}_{\mathbf{t}})|}\sum_{l\in J(\mathbf{x}_{\mathbf{t}})}
\frac{1}{(1-t_l)\,|\mathcal{C}_l(\mathbf{x}_{\mathbf{t}})|}\sum_{i\in \mathcal{C}_l(\mathbf{x}_{\mathbf{t}})}\ell^i\Big],
\label{eq:loss}
\end{equation}
where $\ell^i$ is the loss of Equation~\eqref{eq:entryloss}, and each level is weighted by
$1/(1-t_l)$ as in Equation~\eqref{eq:elbo}.

\subsection{Unmasking strategy}
\label{sec:sampling}

Occupancy codes unmasked in the same step are sampled independently, each from its predicted
probabilities, so they may be inconsistent with each
other~\cite{hayakawa2024distillation,liu2025copula,xu2025energy,besnier2025halton}. Some samplers for
masked generation correct such inconsistencies by remasking unmasked entries, so that these entries are
predicted again~\cite{wang2026remasking}. Our strategy also remasks, and chooses the occupancy codes to
remask by when they were unmasked. Occupancy codes unmasked early are sampled given few unmasked occupancy codes. This holds in
particular for coarse occupancy codes, which are unmasked before most finer occupancy codes exist. Our
strategy therefore remasks the oldest unmasked occupancy codes at each step, so that they are
predicted again given the occupancy codes unmasked after them, including finer occupancy codes in nearby
cells.

Generation starts from $\mathbf{x}_0$ and runs for $N$ steps $k=0,\dots,N-1$. At step $k$, the
target fraction of unmasked occupancy codes at level $l\in J$ is given by the schedule
\begin{equation}
\kappa_l(k)=\min(k/N_l,1)^{\rho},
\label{eq:schedule}
\end{equation}
which reaches $1$ at $k=N_l\le N$. Finer occupancy codes are added as coarser ones are
unmasked, so we let the schedules of coarser levels reach $1$ earlier,
$N_1\le N_3\le\dots\le N_{L-1}=N$.
Early in generation, with few occupancy codes unmasked, the predictions are least certain, so occupancy codes unmasked in the same step are
most likely to be inconsistent with each other. We therefore use $\rho>1$, so that $\kappa_l$
grows slowly at first and quickly later.

Each step $k$ does three things in order. First, for $k\ge1$, each level remasks its oldest unmasked
occupancy codes, so that its fraction of unmasked occupancy codes goes back to about $\kappa_l(k-1)$. When an
occupancy code is remasked, the cells that it marked as occupied are no longer known to be occupied.
As in Section~\ref{sec:process}, the occupancy codes of all cells inside them are then removed from
$\mathbf{x}_{\mathbf{t}}$.
Second, the network predicts all masked occupancy codes in $\mathbf{x}_{\mathbf{t}}$, with each $t_l$ set to the fraction of unmasked occupancy codes at level $l$. Third, each
level unmasks masked occupancy codes chosen at random, so that its fraction rises to about
$\kappa_l(k+1)$. To unmask an occupancy code, we
sample its value from the probabilities predicted in the second step, at temperature $\tau$.
At $\tau=0$, we take the most probable value. Under the schedule, a very large number of occupancy codes at level $L-1$
would be unmasked in step $N-1$, and these would likely be inconsistent with each other. To alleviate this, we add a few extra steps after step $N-1$ and unmask these occupancy codes evenly over step $N-1$ and
the extra steps. The extra steps do not remask. See Section~\ref{sec:impl} in \suppname{} for
details.

\section{Experiments}
\label{sec:exp}

\providecommand{\best}[1]{\textbf{#1}}
\providecommand{\second}[1]{\underline{#1}}

We evaluate our framework on image-conditioned generation (Section~\ref{sec:exp-image}) and
unconditional generation (Section~\ref{sec:exp-shapenet}), measure its efficiency
(Section~\ref{sec:exp-efficiency}), and ablate its designs (Section~\ref{sec:exp-ablation}). All
shapes are represented by their $64^3$ surface occupancy, the finest level of a hierarchy with $L=4$. We also show shapes generated at a higher resolution of $256^3$ by our framework with $L=6$ in Section~\ref{sec:appendix-256} of \suppname{}. We run all baselines in
Sections~\ref{sec:exp-image} and~\ref{sec:exp-shapenet} with their official weights and settings,
and evaluate all methods on the same test data with the same metrics in each task. Details of the data and
metrics are provided in \suppname{}.

\subsection{Image-conditioned generation}
\label{sec:exp-image}

\textbf{Data.} We train on $174{,}528$ assets of TRELLIS-500K~\cite{xiang2025structured} and test
on $1{,}000$ objects of Toys4k~\cite{stojanov2021using}, which are unseen during training. The condition
is a rendered view of each asset.

\textbf{Baselines.} We compare with TRELLIS.2~\cite{xiang2026native}, the state of the art, and
with TRELLIS~\cite{xiang2025structured}, DVD~\cite{xiang2026dvd},
Direct3D-S2~\cite{wu2026direct3d}, TripoSG~\cite{li2025triposg} and
TripoSR~\cite{tochilkin2024triposr}. Voxels are often used to generate meshes, so we evaluate
both the voxels and the meshes of each method. TRELLIS, DVD, TRELLIS.2, our framework and
Direct3D-S2 generate $64^3$ voxels. For the first four, we obtain the meshes by passing the voxels
through the mesh stage of TRELLIS.2, which generates a mesh from $64^3$ voxels. Direct3D-S2 refines its voxels into a mesh with its own second stage, and we evaluate this
mesh. TripoSG and TripoSR generate meshes directly, and their voxel metrics use these meshes
voxelized at $64^3$.

\textbf{Metrics.} We first align each generated shape to its ground truth in orientation and
scale, and then compare it with the ground truth. On the voxels, we report the voxel Chamfer
distance (CD$_\text{V}$) and the voxel IoU, which measure geometric accuracy. A few misplaced
voxels barely change these two metrics but produce visible artifacts, so we also report the LPIPS
(LPIPS$_\text{V}$) and DINOv2~\cite{oquab2023dinov2} similarity (DINO$_\text{V}$) between renderings
of the generated and ground-truth voxels, which measure perceptual similarity. On the meshes, we
report (i) CD$_\text{shell}$, the Chamfer distance between the outer surfaces, which measures
geometric accuracy, (ii) LPIPS$_\text{N}$ and CLIP$_\text{N}$~\cite{radford2021learning} on
rendered normal maps, which measure perceptual similarity, and (iii) the
Uni3D~\cite{zhou2024uni3d} similarity of point clouds sampled on the meshes, which measures semantic
similarity.

\begin{table*}[t]
\centering
\caption{Image-conditioned generation on Toys4k ($^\ast$ voxels from voxelized meshes;
$^\ddagger$ evaluated on their own meshes). The best and second-best values of each column are in bold and underlined, respectively.}
\label{tab:image}
\setlength{\tabcolsep}{4pt}
\resizebox{0.81\textwidth}{!}{%
\begin{tabular}{lcccccccc}
\toprule
& \multicolumn{4}{c}{Voxels} & \multicolumn{4}{c}{Meshes} \\
\cmidrule(lr){2-5}\cmidrule(lr){6-9}
Method & CD$_\text{V}$\,$\downarrow$ & IoU\,$\uparrow$ & LPIPS$_\text{V}$\,$\downarrow$ &
DINO$_\text{V}$\,$\uparrow$ & CD$_\text{shell}$\,($\times10^{-3}$)\,$\downarrow$ &
LPIPS$_\text{N}$\,$\downarrow$ & CLIP$_\text{N}$\,$\uparrow$ &
Uni3D\,$\uparrow$ \\
\midrule
TripoSR$^{\ast\ddagger}$ & 0.01663 & 0.3199 & 0.1837 & 0.7813 & 0.781 & 0.2291 & 0.7431 &
0.3414 \\
TripoSG$^{\ast\ddagger}$ & 0.01325 & 0.3722 & 0.1626 & 0.8523 & 0.525 & 0.1698 & 0.7592 &
0.3849 \\
Direct3D-S2$^\ddagger$ & \second{0.01246} & 0.3694 & 0.1709 & \second{0.8550} & 0.539 &
0.1939 & 0.7600 & 0.3824 \\
DVD & 0.01973 & 0.2786 & 0.2119 & 0.7914 & 1.070 & 0.2412 & 0.7538 & 0.3718 \\
TRELLIS & 0.01328 & 0.3595 & 0.1641 & 0.8385 & \second{0.524} & 0.1899 & 0.7583 &
0.3799 \\
TRELLIS.2 & 0.01250 & \second{0.4017} & \second{0.1524} & 0.8513 & \second{0.524} &
\second{0.1666} & \second{0.7613} & \second{0.3865} \\
Ours & \best{0.01107} & \best{0.4158} & \best{0.1395} & \best{0.8727} & \best{0.407} &
\best{0.1621} & \best{0.7615} & \best{0.3868} \\
\bottomrule
\end{tabular}%
}
\end{table*}

\begin{figure*}[t]
\centering
\includegraphics[width=\linewidth]{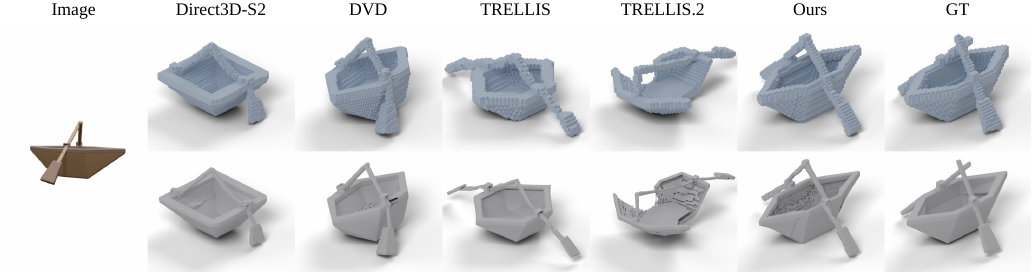}
\caption{An example of image-conditioned generation on Toys4k. The top row shows the generated
$64^3$ voxels and the bottom row the meshes obtained from them. Each shape is aligned to its
ground truth (GT) for display. More examples are in \suppname{}.}
\label{fig:qual}
\end{figure*}

\textbf{Results.} As shown in Table~\ref{tab:image}, our framework achieves the best value on
every voxel metric. Compared with TRELLIS.2, its voxels have an $11\%$ lower Chamfer distance and
an IoU of $0.416$ against $0.402$, and their renderings have the highest perceptual similarity to
the ground truth. Since TRELLIS, DVD, TRELLIS.2 and our framework share one mesh stage, their meshes differ
only in the voxels they start from. Our voxels thus lead to a $22\%$ lower CD$_\text{shell}$ than
those of TRELLIS.2 and to the lowest LPIPS$_\text{N}$ (Figure~\ref{fig:qual}).

\subsection{Unconditional generation}
\label{sec:exp-shapenet}

\textbf{Data.} We use the chair, airplane, table and rifle categories of
ShapeNet~\cite{chang2015shapenet}, with the training and test splits of
LAS-Diffusion~\cite{zheng2023lasdiffusion}. The test split of each category,
voxelized at $64^3$, serves as its reference set.

\textbf{Baselines.} We compare with LAS-Diffusion, OctFusion~\cite{xiong2025octfusion},
OctGPT~\cite{wei2025octgpt} and XCube~\cite{ren2024xcube}. XCube releases models only for chair
and airplane, so we compare with it on these two categories. LAS-Diffusion generates $64^3$
occupancy, and XCube generates a sparse $512^3$ occupancy, which we max-pool to $64^3$.
OctFusion and OctGPT output a $32^3$ occupancy and a mesh. We evaluate both outputs of these two
methods, voxelizing the meshes at $64^3$ in the same way as the reference sets. Rows that use the
$32^3$ occupancy are marked in the tables.

\textbf{Metrics.} Unconditional samples have no corresponding ground truth, so we compare the
generated set of each category with its reference set. We report the 1-NNA, the coverage (COV)
and the minimum matching distance (MMD)~\cite{achlioptas2018learning,yang2019pointflow}, which measure distribution
similarity, diversity and fidelity, respectively, all computed with the Chamfer distance between
voxel centers. We also report FID$_\text{V}$ on PointNet++ features of Point-E~\cite{nichol2022point},
and FID$_\text{VD}$ and KID$_\text{VD}$ on DINOv2 features of voxel renderings, which measure the
distance between the generated and reference distributions in a 3D feature space and an image
feature space, respectively.

\begin{table}[t]
\centering
\caption{Unconditional generation on ShapeNet, averaged over the four categories, and over chair
and airplane for the last two rows. For 1-NNA, closer to $0.5$ is better. MMD is in units of
$10^{-3}$ and KID$_\text{VD}$ in units of $10^{-2}$ ($^\dagger$ $32^3$ occupancy output; ``mesh''
meshes voxelized at $64^3$). The best and second-best values within each block are in bold and underlined, respectively.}
\label{tab:shapenet}
\setlength{\tabcolsep}{3pt}
\resizebox{\columnwidth}{!}{%
\begin{tabular}{lcccccc}
\toprule
Method & 1-NNA & COV\,$\uparrow$ & MMD\,$\downarrow$ & FID$_\text{V}$\,$\downarrow$ &
FID$_\text{VD}$\,$\downarrow$ & KID$_\text{VD}$\,$\downarrow$ \\
\midrule
\multicolumn{7}{l}{\emph{Four categories}} \\
LAS-Diffusion & 0.782 & 0.365 & \second{2.08} & 63.3 & 82.5 & 16.7 \\
OctFusion$^\dagger$ & 0.805 & 0.290 & 2.23 & 46.2 & 184.5 & 41.5 \\
OctFusion (mesh) & 0.779 & 0.310 & 2.23 & 11.1 & \second{43.0} & \second{4.99} \\
OctGPT$^\dagger$ & 0.773 & 0.388 & 2.22 & 43.2 & 186.6 & 42.5 \\
OctGPT (mesh) & \second{0.736} & \second{0.412} & 2.25 & \second{10.1} & 61.0 &
10.4 \\
Ours & \best{0.574} & \best{0.501} & \best{1.91} & \best{2.06} & \best{15.3} & \best{0.56} \\
\midrule
\multicolumn{7}{l}{\emph{Chair and airplane}} \\
XCube & 0.615 & 0.485 & \best{1.89} & 4.74 & 20.8 & 1.91 \\
Ours & \best{0.576} & \best{0.502} & 2.05 & \best{1.79} & \best{15.3} & \best{0.54} \\
\bottomrule
\end{tabular}%
}
\end{table}

\begin{figure*}[t]
\vspace{-2em}
\centering
\includegraphics[width=\linewidth]{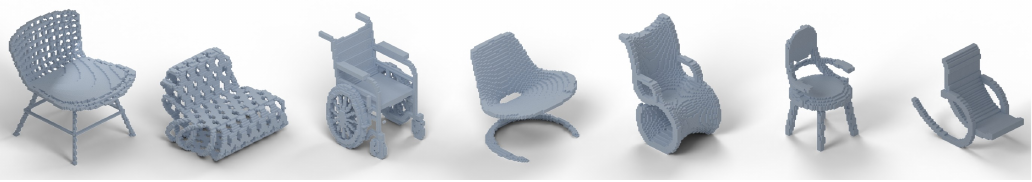}
\caption{Unconditional generation of chairs by our framework on ShapeNet.}
\label{fig:chairs}
\vspace{-1em}
\end{figure*}

\textbf{Results.} As shown in Table~\ref{tab:shapenet}, our framework achieves the best value on
all six metrics, averaged over the four categories. Its FID$_\text{V}$ is $80\%$ lower and its
FID$_\text{VD}$ is $64\%$ lower than the best baseline values. OctFusion, OctGPT and XCube are
hierarchical methods, whose coarse levels are generated with the least information about the
shape and are never revisited. Our framework learns the joint distribution of all levels and
outperforms OctFusion and OctGPT on the four categories and XCube on chair and airplane.
Table~\ref{tab:shapenet-cat} in \suppname{} gives the results of each category.
Figure~\ref{fig:chairs} shows chairs generated by our framework, with fine and complex structures.

\subsection{Efficiency}
\label{sec:exp-efficiency}

\begin{figure}[t]
\centering
\includegraphics[width=\linewidth]{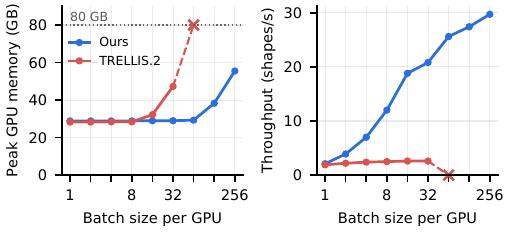}
\caption{Peak GPU memory and training throughput of our network and of the voxel stage of
TRELLIS.2, with the same backbone, at each batch size on one A100 GPU with 80\,GB. A cross marks
the batch size at which TRELLIS.2 runs out of memory.}
\label{fig:efficiency}
\end{figure}

We compare one training step (a forward pass, a backward pass and an optimizer update) of our
network and of the voxel stage of TRELLIS.2 with the same backbone. Both run on one A100 GPU with 80\,GB, with gradient checkpointing, fp32 master weights and
bf16 mixed precision. As shown in Figure~\ref{fig:efficiency}, at a batch size of $32$ our
framework needs $29.0$\,GB of GPU memory instead of $47.3$\,GB and has $8\times$ higher training
throughput, and it can be trained with batch sizes at which TRELLIS.2 runs out of memory.
Since
the two networks share the backbone, these differences in memory and throughput come from
the number of cells they process. TRELLIS.2 processes all $4{,}096$ cells of its $16^3$ latent grid for every shape,
including those in empty space, while our network processes only the cells that have a token.
In a training state $\mathbf{x}_{\mathbf{t}}$, which is partly masked, $273$ cells have a token on
average. A complete shape, in which no occupancy code is masked as at the end of generation, has $579$
on average over the Toys4k test set. One forward pass of our network therefore takes on average about $1/15$
of the FLOPs of the voxel stage of TRELLIS.2 ($0.82$ vs.\ $12.2$\,TFLOPs).

To measure the cost of full training, we train our framework on four H100 GPUs with a batch size of $256$. TRELLIS and TRELLIS.2 are trained with the same batch size on $64$ A100 GPUs and $32$ H100 GPUs, respectively, $16$ and $8$ times as many GPUs as we use. After about two days of training ($80$K steps), our framework
surpasses TRELLIS, which is trained for $400$K steps, on every metric of Section~\ref{sec:exp-image}. After about two
and a half days, it matches TRELLIS.2 (Table~\ref{tab:train-time} in \suppname{}), whose paper does not report the number of training steps. Our framework also generates the voxels faster than the voxel stage
of TRELLIS.2 (Table~\ref{tab:infer-time} in \suppname{}).
\subsection{Ablation studies}
\label{sec:exp-ablation}

We conduct ablation studies on image-conditioned generation on Toys4k.

\begin{table}[t]
\centering
\caption{Ablation of two-level prediction and joint hierarchical modeling on Toys4k. The first
two variants generate the hierarchy level by level. CD$_\text{shell}$ is in units of $10^{-3}$. The best value of each column is in bold.}
\label{tab:ablation-design}
\setlength{\tabcolsep}{3pt}
\resizebox{0.88\columnwidth}{!}{%
\begin{tabular}{lccccc}
\toprule
Variant & CD$_\text{V}$\,$\downarrow$ & IoU\,$\uparrow$ & CD$_\text{shell}$\,$\downarrow$ &
LPIPS$_\text{N}$\,$\downarrow$ & Uni3D\,$\uparrow$ \\
\midrule
Single-level & 0.02150 & 0.238 & 1.280 & 0.2566 & 0.3508 \\
Two-level & 0.01941 & 0.264 & 1.096 & 0.2428 & 0.3575 \\
Ours & \best{0.01506} & \best{0.334} & \best{0.744} & \best{0.2019} &
\best{0.3820} \\
\bottomrule
\end{tabular}%
}
\end{table}

\textbf{Two-level prediction and joint hierarchical modeling.} We compare our framework with
two variants in Table~\ref{tab:ablation-design}. The single-level variant uses single-level prediction
(Section~\ref{sec:twolevel}) and the two-level variant uses two-level prediction, and both
generate the hierarchy level by level. By contrast, ours extends the two-level variant with \emph{joint} hierarchical modeling.
All three share the network architecture and the training
data, and are trained for the same number of steps, fewer than those used for our model in
Section~\ref{sec:exp-image}. As shown in Table~\ref{tab:ablation-design}, the two-level variant
improves every metric over the single-level variant. Compared with the two-level variant, our
framework lowers the voxel Chamfer distance by $22\%$ and CD$_\text{shell}$ by $32\%$, and
improves every metric. In addition, two-level prediction makes training faster, since the
network processes fewer tokens. With both variants on four A100 GPUs at a batch size of $256$, one
training step of the two-level variant takes $4.03$\,s, compared with $8.96$\,s for the
single-level variant, a $2.2\times$ speedup.

\begin{table}[t]
\centering
\caption{Ablation of the unmasking strategy on Toys4k. The best and
second-best values of each column are in bold and underlined, respectively.}
\label{tab:ablation-sampling}
\small
\setlength{\tabcolsep}{10pt}
\begin{tabular}{lcc}
\toprule
Sampler & CD$_\text{V}$\,$\downarrow$ & IoU\,$\uparrow$ \\
\midrule
MDLM~\cite{sahoo2024simple} & 0.01162 & 0.4035 \\
ReMDM~\cite{wang2026remasking} & \second{0.01115} & 0.4085 \\
ReMDM-conf~\cite{wang2026remasking} & 0.01116 & \second{0.4139} \\
MaskGIT~\cite{chang2022maskgit} & 0.01243 & 0.3987 \\
\midrule
Ours & \best{0.01107} & \best{0.4158} \\
\bottomrule
\end{tabular}
\end{table}

\textbf{Unmasking strategy.} We compare our unmasking strategy with other samplers, all applied
to our model of Section~\ref{sec:exp-image}. MDLM~\cite{sahoo2024simple} does not remask.
ReMDM~\cite{wang2026remasking} remasks unmasked occupancy codes chosen at random, and its
confidence-based variant ReMDM-conf~\cite{wang2026remasking} remasks the less confident ones with
a higher probability. MaskGIT~\cite{chang2022maskgit} unmasks the most probable occupancy codes first
and does not remask. Unlike these samplers, our strategy chooses the occupancy codes to remask by when they were unmasked. An occupancy code unmasked earlier is predicted given fewer finer occupancy codes in nearby cells, so its prediction is less certain. Our
strategy therefore remasks the oldest unmasked occupancy codes, so that they are revisited once
finer occupancy codes are unmasked. As shown in
Table~\ref{tab:ablation-sampling}, our strategy gives the lowest voxel Chamfer distance and the
highest IoU among all samplers.

\section{Limitations and future work}

A limitation of our model is the size of its training set. Due to limited training resources,
we train it on about $175$K assets, less than a quarter of the roughly $800$K assets used to train
TRELLIS.2~\cite{xiang2026native}. While our model, with the data augmentation described in \suppname{}, still
outperforms TRELLIS.2 in generalization to the unseen objects of
Toys4k, we expect more training data to further improve it. Besides,
like other masked discrete flow matching models, our model unmasks multiple occupancy codes independently in one step,
which can make them inconsistent with each
other~\cite{hayakawa2024distillation,liu2025copula,xu2025energy,besnier2025halton}. Our unmasking
strategy is designed to reduce such inconsistencies, and exploring better unmasking strategies
may further improve the quality of the generated voxels.

In future work, we plan to apply our framework to generating large scenes with
multiple objects and long sequences of 3D shapes. Both contain far more voxels than a single
object, and our framework keeps their training lightweight by processing only the cells inside
occupied cells.

\section{Conclusion}

We presented JoVox, a discrete flow matching framework that learns the joint distribution of all
levels of the voxel hierarchy, so that finer cells take part in deciding coarse cells. Within this framework,
two-level prediction halves the number of levels that the network processes and yields voxels of
higher fidelity, and our unmasking strategy revisits early decisions once finer cells are
generated. Our framework outperforms the baselines on unconditional and image-conditioned
generation, while its training is far cheaper than that of TRELLIS.2. \textbf{Our results
thus show that voxels can be generated with both clearly improved fidelity and drastically improved efficiency.}

{
    \small
    \bibliographystyle{ieeenat_fullname}
    \bibliography{main}
}

\ifdefined\standalonesupp\else\clearpage\fi %
\setcounter{section}{0}
\renewcommand{\thesection}{\Alph{section}}

\section{Generation at $256^3$}
\label{sec:appendix-256}

\begin{figure*}[t]
\centering
\includegraphics[width=\linewidth]{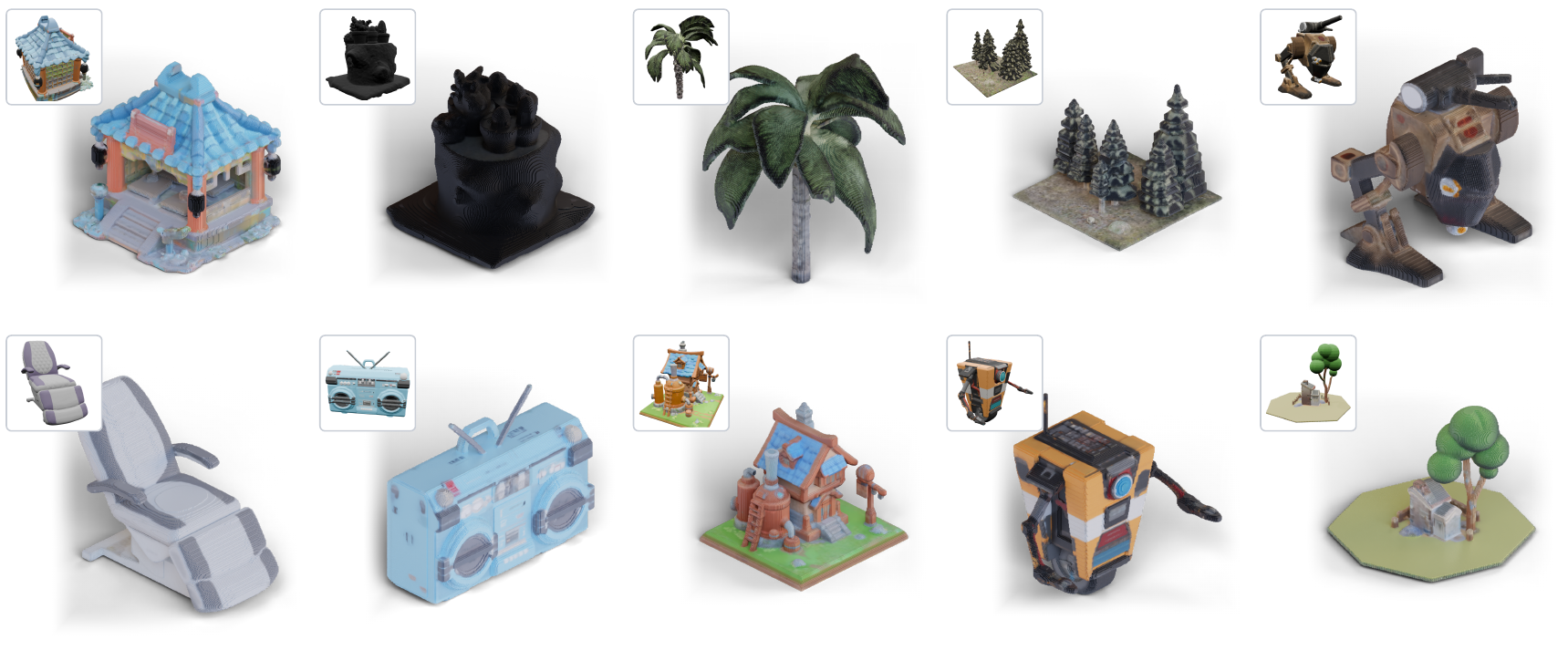}
\caption{Image-conditioned generation at $256^3$. Each input image is shown in the box at the top
left of the generated $256^3$ voxels, which are colored by the texture stage of TRELLIS.2.}
\label{fig:gen256}
\end{figure*}

Figure~\ref{fig:gen256} shows shapes generated at $256^3$ from input images by our framework with
$L=6$ (Section~\ref{sec:impl}).

\section{Implementation details}
\label{sec:impl}

The experiments in Section~\ref{sec:exp} use $L=4$, so the finest level is a grid of $64^3$ cells. The tokens come from
the cells at level $0$ and the occupied cells at level $2$, with grids of $4^3$ and $16^3$ cells, and the occupancy codes
are at levels $1$ and $3$.
The $256^3$ shapes in Section~\ref{sec:appendix-256} and on the \webname{} are generated by a model with $L=6$, where the
tokens come from the cells at level $0$ and the occupied cells at levels $2$ and $4$, and the occupancy codes are at levels $1$, $3$
and $5$. For this model, we add the level embedding $\mathrm{emb}_{\text{level}}(4)$ and the
embedding $\mathrm{emb}_t^{\,5}$ of $t_5$, initialize all other weights from the model with $L=4$, and train it
for $32$K further steps on four GPUs.

The transformer has $1.54$B parameters in $30$ blocks of width $1536$ with $12$ heads. The
image encoder is a frozen DINOv3 ViT-L/16 at a resolution of $512\times512$. The condition
is dropped with probability $0.1$ during training, and classifier-free guidance at scale $1.5$
is applied to the logits during sampling.

We train with AdamW~\cite{loshchilov2017decoupled} at a learning rate of $10^{-4}$ and a
batch size of $256$ on four GPUs, and keep an exponential moving average of the weights. For data
augmentation, the condition image is randomly perturbed in brightness, saturation, contrast, hue
and gamma, and the occupancy is rotated by a random multiple of $90^\circ$ about the vertical
axis. We draw $t_l=v_l^{2}$ with $v_l\sim\mathcal{U}(0,1)$.

In the unmasking strategy of Section~\ref{sec:sampling}, we remask at the levels one after another at each step, from the coarsest to the finest, so that we count the unmasked occupancy codes at a finer level only after remasking at the coarser levels has removed some of them. At step $k\ge1$, at each level $l$ with $\kappa_l(k)<1$, we remask up to $\operatorname{round}\big((1-\kappa_l(k-1)/\kappa_l(k))\,n\big)$ occupancy codes, rounded half to even, where $n$ is the number of all unmasked occupancy codes at level $l$. We do not remask an occupancy code unmasked at step $k-1$, since no prediction has yet been made given it, nor an occupancy code unmasked in a step in which no other occupancy code, at any level, was unmasked. Among the other unmasked occupancy codes, we remask those last unmasked earliest, with ties broken at random, or all of them if there are fewer. For the prediction, we set $t_l$ to $0$ at a level that has no occupancy codes. After the prediction, we unmask only occupancy codes predicted at this step, and occupancy codes added after an unmasking are first predicted at the next step and are not counted below. At a level $l$ with $\kappa_l(k-1)=1$, we unmask all of its masked occupancy codes. At any other level $l$ with masked occupancy codes, we unmask the proportion $\big(\kappa_l(k+1)-\kappa_l(k-1)\big)/\big(1-\kappa_l(k-1)\big)$ of them, rounded up and at least one, chosen at random, where $\kappa_l(-1)=0$. From step $N-1$ on, we instead unmask the proportion $1/(N+N_\text{f}-k)$ of the masked occupancy codes at level $L-1$ at step $k$, also rounded up and at least one, so that they are unmasked evenly over step $N-1$ and $N_\text{f}$ extra steps, and we do not remask in the extra steps.

For image-conditioned generation, the sampler uses $N=72$,
$N_\text{f}=8$, $N_1=0.8N$, $\rho=2$ and $\tau=0$, so a generation takes $80$ steps. For
unconditional generation, it uses $\tau=1$, with $N=288$ for chair and table and
$N=72$ for airplane and rifle.

Table~\ref{tab:infer-time} reports the time and the peak GPU memory of generating the voxels.
Our framework generates the voxels faster than the voxel stage of TRELLIS.2 at every batch size.

\begin{table}[t]
\centering
\caption{Time per shape and peak GPU memory of generating the voxels for image-conditioned
generation on one A100 GPU with 80\,GB, for our framework and the voxel stage of TRELLIS.2 (OOM,
out of memory). The time includes image encoding, with $80$ sampling steps for our framework and the
default $12$ for TRELLIS.2.}
\label{tab:infer-time}
\setlength{\tabcolsep}{4pt}
\begin{tabular}{rcccc}
\toprule
& \multicolumn{2}{c}{Ours} & \multicolumn{2}{c}{TRELLIS.2} \\
\cmidrule(lr){2-3}\cmidrule(lr){4-5}
Batch & Time (s) & Memory (GB) & Time (s) & Memory (GB) \\
\midrule
$8$ & $2.02$ & $6.1$ & $2.30$ & $5.7$ \\
$16$ & $1.77$ & $7.9$ & $2.24$ & $11.2$ \\
$32$ & $1.80$ & $11.3$ & $2.21$ & $22.2$ \\
$64$ & $1.67$ & $17.8$ & $2.20$ & $44.2$ \\
$128$ & $1.62$ & $28.9$ & \multicolumn{2}{c}{OOM} \\
\bottomrule
\end{tabular}
\end{table}

\begin{table*}[t]
\centering
\caption{Unconditional generation on each ShapeNet category (test splits). For 1-NNA, closer
to $0.5$ is better. MMD is in units of $10^{-3}$ and KID$_\text{VD}$ in units of $10^{-2}$.
Rows marked $^\dagger$ use the $32^3$ occupancy the method outputs, and rows marked ``mesh''
use the method's meshes voxelized at $64^3$. XCube releases models only for chair and airplane. The best and second-best values
of each column within each category are in bold and underlined, respectively.}
\label{tab:shapenet-cat}
\setlength{\tabcolsep}{3pt}
\resizebox{\textwidth}{!}{%
\begin{tabular}{lcccccccccccc}
\toprule
& \multicolumn{6}{c}{Chair} & \multicolumn{6}{c}{Airplane} \\
\cmidrule(lr){2-7}\cmidrule(lr){8-13}
Method & 1-NNA & COV\,$\uparrow$ & MMD\,$\downarrow$ & FID$_\text{V}$\,$\downarrow$ &
FID$_\text{VD}$\,$\downarrow$ & KID$_\text{VD}$\,$\downarrow$ & 1-NNA & COV\,$\uparrow$ &
MMD\,$\downarrow$ & FID$_\text{V}$\,$\downarrow$ & FID$_\text{VD}$\,$\downarrow$ &
KID$_\text{VD}$\,$\downarrow$ \\
\midrule
LAS-Diffusion & 0.697 & 0.432 & 3.40 & 30.67 & 57.6 & 8.77 &
0.881 & 0.357 & 1.06 & 73.07 & 130.3 & 28.54 \\
OctFusion$^\dagger$ & 0.689 & 0.406 & 3.59 & 22.49 & 106.8 & 15.63 &
0.902 & 0.261 & 1.17 & 51.04 & 305.2 & 67.86 \\
OctFusion (mesh) & 0.671 & 0.412 & 3.68 & 5.01 & 29.5 & \second{2.70} &
0.863 & 0.268 & 1.00 & 12.64 & 51.2 & 6.41 \\
OctGPT$^\dagger$ & 0.715 & 0.397 & 3.74 & 25.38 & 139.6 & 26.30 &
0.872 & 0.365 & 1.09 & 47.83 & 312.3 & 68.88 \\
OctGPT (mesh) & 0.697 & 0.401 & 3.90 & 9.36 & 77.8 & 14.74 &
0.821 & 0.361 & 0.93 & 12.78 & 27.9 & 2.21 \\
XCube & \best{0.541} & \second{0.503} & \best{3.03} & \second{2.85} & \second{24.9} & 2.74 &
\second{0.689} & \second{0.467} & \best{0.75} & \second{6.62} & \second{16.6} & \second{1.07} \\
Ours & \second{0.597} & \best{0.504} & \second{3.31} & \best{2.01} & \best{19.1} & \best{0.88} &
\best{0.554} & \best{0.499} & \second{0.78} & \best{1.57} & \best{11.5} & \best{0.20} \\
\midrule
& \multicolumn{6}{c}{Table} & \multicolumn{6}{c}{Rifle} \\
\cmidrule(lr){2-7}\cmidrule(lr){8-13}
LAS-Diffusion & 0.727 & 0.401 & \second{2.92} & 58.25 & 80.5 & 14.43 &
0.823 & 0.269 & 0.95 & 91.18 & 61.4 & 15.20 \\
OctFusion$^\dagger$ & 0.788 & 0.312 & 3.13 & 34.45 & 86.5 & 10.35 &
0.839 & 0.181 & 1.02 & 76.75 & 239.4 & 72.30 \\
OctFusion (mesh) & 0.769 & 0.327 & 3.19 & 12.39 & \second{64.3} & \second{7.24} &
0.813 & 0.232 & 1.06 & 14.20 & 27.1 & 3.61 \\
OctGPT$^\dagger$ & 0.724 & 0.451 & 3.15 & 35.21 & 90.1 & 12.58 &
0.780 & 0.339 & \second{0.88} & 64.31 & 204.3 & 62.07 \\
OctGPT (mesh) & \second{0.684} & \second{0.480} & 3.24 & \second{10.65} & 118.1 & 21.86 &
\second{0.742} & \second{0.406} & 0.92 & \second{7.67} & \second{20.2} & \second{2.60} \\
Ours & \best{0.562} & \best{0.520} & \best{2.73} & \best{1.73} & \best{18.4} & \best{0.83} &
\best{0.581} & \best{0.482} & \best{0.83} & \best{2.92} & \best{12.0} & \best{0.32} \\
\bottomrule
\end{tabular}%
}
\end{table*}

\section{Data and evaluation details}
\label{sec:appendix-eval}

\textbf{Data.} The $174{,}528$ training assets come from the Sketchfab, ABO, 3D-FUTURE and HSSD
parts of TRELLIS-500K~\cite{xiang2025structured}, about $35\%$ of the full dataset. The
condition of a training asset is a rendered view at an elevation between $-10^\circ$ and
$30^\circ$, and each Toys4k test object has one condition view at a similar elevation. On
ShapeNet, each category is evaluated separately on its test split, and its reference set has
as many shapes as the generated set.

\textbf{Alignment.} Methods and datasets differ in their orientation and scale conventions.
We therefore align each generated shape to its ground truth by picking the one of the $24$
axis-aligned rotations that minimizes the voxel Chamfer distance and refining it with a scale-aware ICP.

\textbf{Voxel metrics.} For LPIPS$_\text{V}$
and DINO$_\text{V}$, we render the aligned voxels and the ground-truth voxels from six views and
average the LPIPS and the cosine similarity of DINOv2 features over the views. For Direct3D-S2,
the voxel metrics use the $64^3$ voxels of its dense first stage, and for TripoSG and TripoSR, they use their
voxelized meshes.

\textbf{Mesh metrics.} The meshes of Direct3D-S2, TripoSG and TripoSR are aligned to the ground
truth in the same way. CD$_\text{shell}$ is the Chamfer distance between the outer surfaces of
the two meshes. LPIPS$_\text{N}$ and CLIP$_\text{N}$ compare normal maps rendered from four
views. Uni3D is the similarity between the Uni3D embeddings of
point clouds sampled on the two meshes.

\textbf{ShapeNet metrics.} FID$_\text{V}$ is computed on points of the voxels. FID$_\text{VD}$
and KID$_\text{VD}$ use renderings of the voxels from six views.

\section{Additional ablations}
\label{sec:appendix-training}

This section measures two choices of Section~\ref{sec:impl} on the first $200$ objects of
Toys4k, with the unmasking strategy of Section~\ref{sec:sampling}.

\textbf{Exponential moving average of the weights.} For our model, the moving average improves
CD$_\text{V}$ from $0.01215$ to $0.01193$ and CD$_\text{shell}$ from $0.503\times10^{-3}$ to
$0.446\times10^{-3}$. It also improves LPIPS$_\text{N}$ from $0.1758$ to $0.1714$.

\textbf{Augmentation.} We train the same model with and without the augmentation of
Section~\ref{sec:impl} and compare the two after the same number of training steps
(Table~\ref{tab:aug}). Augmentation improves CD$_\text{V}$, CD$_\text{shell}$, LPIPS$_\text{N}$,
CLIP$_\text{N}$ and Uni3D.

\begin{table}[h]
\centering
\caption{Training with and without augmentation, compared after the same number of training steps
on the first $200$ objects of Toys4k. CD$_\text{shell}$ is in units of $10^{-3}$. The better value
of each column is in bold.}
\label{tab:aug}
\setlength{\tabcolsep}{3pt}
\resizebox{0.88\columnwidth}{!}{%
\begin{tabular}{lcccccc}
\toprule
Aug. & CD$_\text{V}$\,$\downarrow$ & IoU\,$\uparrow$ & CD$_\text{shell}$\,$\downarrow$ &
LPIPS$_\text{N}$\,$\downarrow$ & CLIP$_\text{N}$\,$\uparrow$ &
Uni3D\,$\uparrow$ \\
\midrule
 & 0.01280 & \best{0.3923} & 0.568 & 0.1858 & 0.7588 & 0.3844 \\
\checkmark & \best{0.01271} & 0.3790 & \best{0.524} & \best{0.1797} & \best{0.7608} &
\best{0.3847} \\
\bottomrule
\end{tabular}%
}
\end{table}

\section{Per-category results on ShapeNet}
\label{sec:appendix-shapenet}

Table~\ref{tab:shapenet-cat} gives the results of Table~\ref{tab:shapenet} in the main paper for each
category.

\section{Results after shorter training}
\label{sec:appendix-time}

Table~\ref{tab:train-time} compares our framework at the two training lengths of
Section~\ref{sec:exp-efficiency} with TRELLIS and TRELLIS.2 on Toys4k.

\begin{table*}[t]
\centering
\caption{Our framework after about two days and two and a half days of training on four H100 GPUs,
compared with TRELLIS and TRELLIS.2 on Toys4k. Each of our rows is compared with the baseline above
it. Steps are counted in batches of $256$ shapes.}
\label{tab:train-time}
\setlength{\tabcolsep}{4pt}
\resizebox{0.94\textwidth}{!}{%
\begin{tabular}{lcccccccc}
\toprule
& \multicolumn{4}{c}{Voxels} & \multicolumn{4}{c}{Meshes} \\
\cmidrule(lr){2-5}\cmidrule(lr){6-9}
Method & CD$_\text{V}$\,$\downarrow$ & IoU\,$\uparrow$ & LPIPS$_\text{V}$\,$\downarrow$ &
DINO$_\text{V}$\,$\uparrow$ & CD$_\text{shell}$\,($\times10^{-3}$)\,$\downarrow$ &
LPIPS$_\text{N}$\,$\downarrow$ & CLIP$_\text{N}$\,$\uparrow$ &
Uni3D\,$\uparrow$ \\
\midrule
TRELLIS & 0.01328 & 0.3595 & 0.1641 & 0.8385 & 0.524 & 0.1899 & 0.7583 & 0.3799 \\
Ours, about two days ($80$K steps) & 0.01221 & 0.3893 & 0.1485 & 0.8614 & 0.497 & 0.1737 &
0.7600 & 0.3851 \\
\midrule
TRELLIS.2 & 0.01250 & 0.4017 & 0.1524 & 0.8513 & 0.524 & 0.1666 & 0.7613 & 0.3865 \\
Ours, about two and a half days ($104$K steps) & 0.01137 & 0.4035 & 0.1424 & 0.8692 & 0.421 &
0.1668 & 0.7612 & 0.3859 \\
\bottomrule
\end{tabular}%
}
\end{table*}

\section{Qualitative comparison}
\label{sec:qual-more}

\begin{figure*}[t]
\centering
\includegraphics[width=\linewidth]{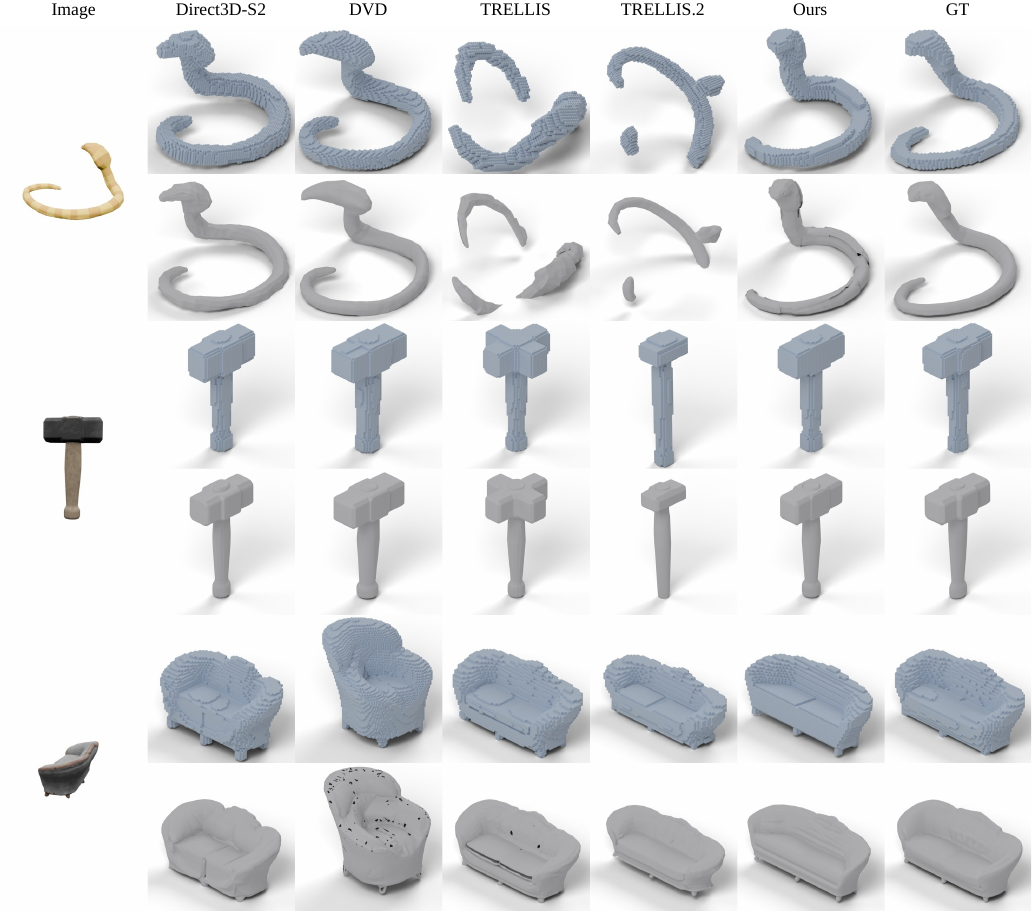}
\caption{Image-conditioned generation on Toys4k. For each input image, the top row shows the
generated $64^3$ voxels and the bottom row the meshes obtained from them as in
Section~\ref{sec:exp-image}. The ground truth (GT) shows the voxels and the mesh of the shape.
Each shape is aligned to its ground truth for display.}
\label{fig:qual-more-a}
\end{figure*}

\begin{figure*}[t]
\centering
\includegraphics[width=\linewidth]{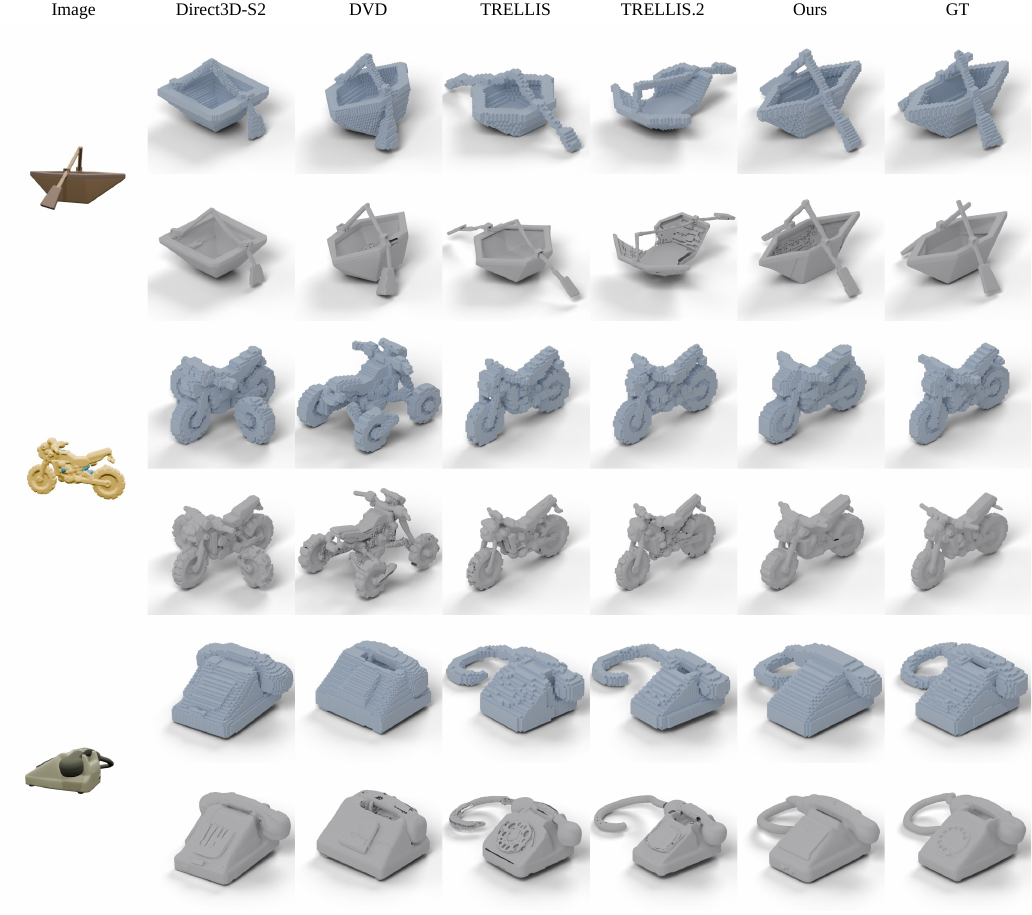}
\caption{Image-conditioned generation on Toys4k, continued from Figure~\ref{fig:qual-more-a}.}
\label{fig:qual-more-b}
\end{figure*}

\begin{figure*}[t]
\centering
\includegraphics[width=\linewidth]{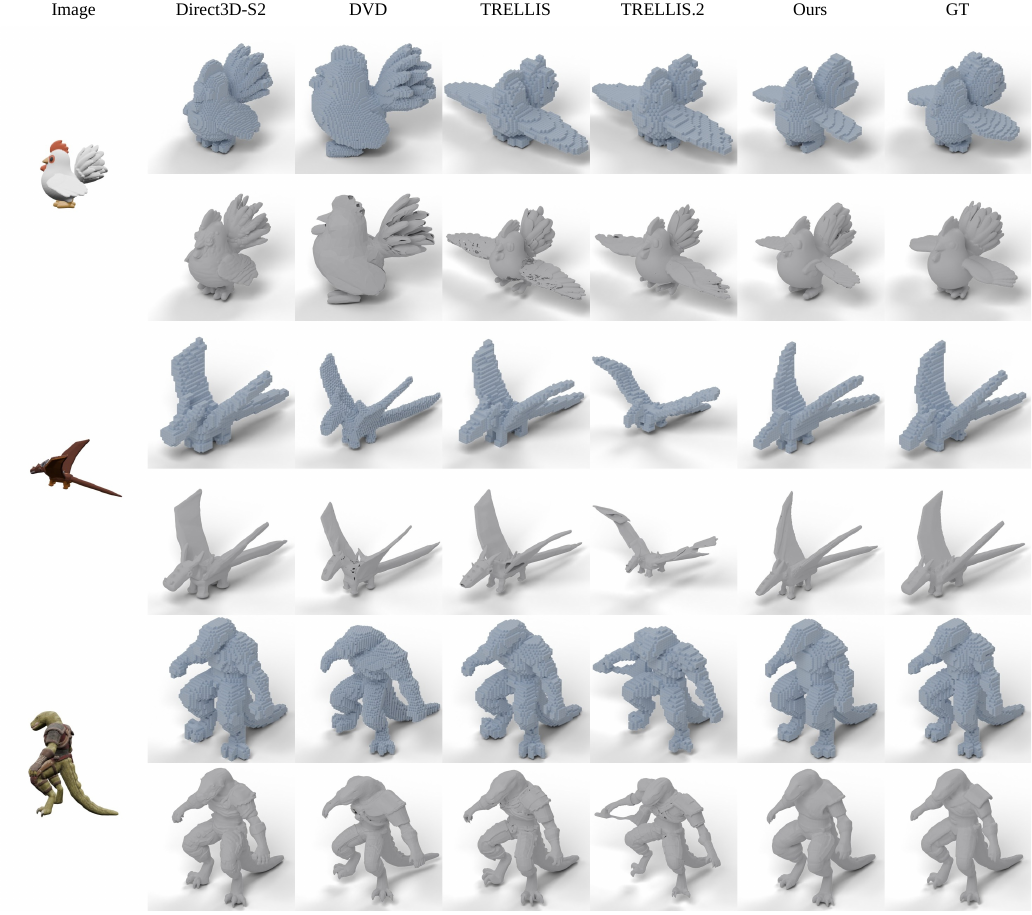}
\caption{Image-conditioned generation on Toys4k, continued from Figure~\ref{fig:qual-more-b}.}
\label{fig:qual-more-c}
\end{figure*}

Figures~\ref{fig:qual-more-a}, \ref{fig:qual-more-b} and~\ref{fig:qual-more-c} compare the
shapes generated from the same image by each method on Toys4k.

\end{document}